\documentclass[10pt,twocolumn,letterpaper]{article}

\usepackage{cvpr}              %

\usepackage{graphicx}
\usepackage{amsmath}
\usepackage{amssymb}
\usepackage{booktabs}
\usepackage{adjustbox}

\usepackage{color}

\usepackage[normalem]{ulem}
\useunder{\uline}{\ul}{}

\usepackage[colorlinks]{hyperref}

\usepackage[capitalize]{cleveref}
\crefname{section}{Sec.}{Secs.}
\Crefname{section}{Section}{Sections}
\Crefname{table}{Table}{Tables}
\crefname{table}{Tab.}{Tabs.}

\definecolor{somegray}{rgb}{0.5, 0.5, 0.5}
\newcommand{\darkgrayed}[1]{\textcolor{somegray}{#1}}
\makeatletter
\newcommand*\titleheader[1]{\gdef\@titleheader{#1}}
\AtBeginDocument{%
  \let\st@red@title\@title
  \def\@title{%
    \vskip-4em
    \bgroup\normalfont\large\centering\@titleheader\par\egroup
    \vskip1.5em\st@red@title}
}
\makeatother

\titleheader{\darkgrayed{This paper has been accepted for publication at the\\
IEEE Conference on Computer Vision and Pattern Recognition (CVPR), Vancouver, 2023.
\copyright IEEE}}

\title{Recurrent Vision Transformers for Object Detection with Event Cameras}

\author{Mathias Gehrig and Davide Scaramuzza\\[5pt]
Robotics and Perception Group, University of Zurich, Switzerland
}

\begin{document}
\maketitle

\begin{abstract}
We present Recurrent Vision Transformers (RVTs), a novel backbone for object detection with event cameras.
Event cameras provide visual information with sub-millisecond latency at a high-dynamic range and with strong robustness against motion blur.
These unique properties offer great potential for low-latency object detection and tracking in time-critical scenarios.
Prior work in event-based vision has achieved outstanding detection performance but at the cost of substantial inference time, typically beyond 40 milliseconds.
By revisiting the high-level design of recurrent vision backbones, we reduce inference time by a factor of 6 while retaining similar performance.
To achieve this, we explore a multi-stage design that utilizes three key concepts in each stage:
First, a convolutional prior that can be regarded as a conditional positional embedding.
Second, local and dilated global self-attention for spatial feature interaction.
Third, recurrent temporal feature aggregation to minimize latency while retaining temporal information.
RVTs can be trained from scratch to reach state-of-the-art performance on event-based object detection - achieving an mAP of 47.2\% on the Gen1 automotive dataset. At the same time, RVTs offer fast inference ($<12$ ms on a T4 GPU) and favorable parameter efficiency ($5\times$ fewer than prior art).
Our study brings new insights into effective design choices that can be fruitful for research beyond event-based vision.
\end{abstract}
\\[-0.3cm]
\noindent
\textbf{Code}: \url{https://github.com/uzh-rpg/RVT}
\section{Introduction}\label{sec:intro}

Time matters for object detection.
In 30 milliseconds, a human can run 0.3 meters, a car on public roads covers up to 1 meter, and a train can travel over 2 meters.
Yet, during this time, an ordinary camera captures only a single frame.

Frame-based sensors must strike a balance between latency and bandwidth.
Given a fixed bandwidth, a frame-based camera must trade-off camera resolution and frame rate.
However, in highly dynamic scenes, reducing the resolution or the frame rate may come at the cost of missing essential scene details, and, in safety-critical scenarios like automotive, this may even cause fatalities.

In recent years, event cameras have emerged as alternative sensor that offers a different trade-off.
Instead of counterbalancing bandwidth requirements and perceptual latency, they provide visual information at sub-millisecond latency but sacrifice absolute intensity information.
Instead of capturing intensity images, event cameras measure changes in intensity at the time they occur.
This results in a stream of events, which encode time, location, and polarity of brightness changes \cite{Gallego20pami}.
The main advantages of event cameras are their sub-millisecond latency, very high dynamic range ($>120$ dB), strong robustness to motion blur, and ability to provide events asynchronously in a continuous manner.

\begin{figure}
    \centering
    \includegraphics[width=\linewidth]{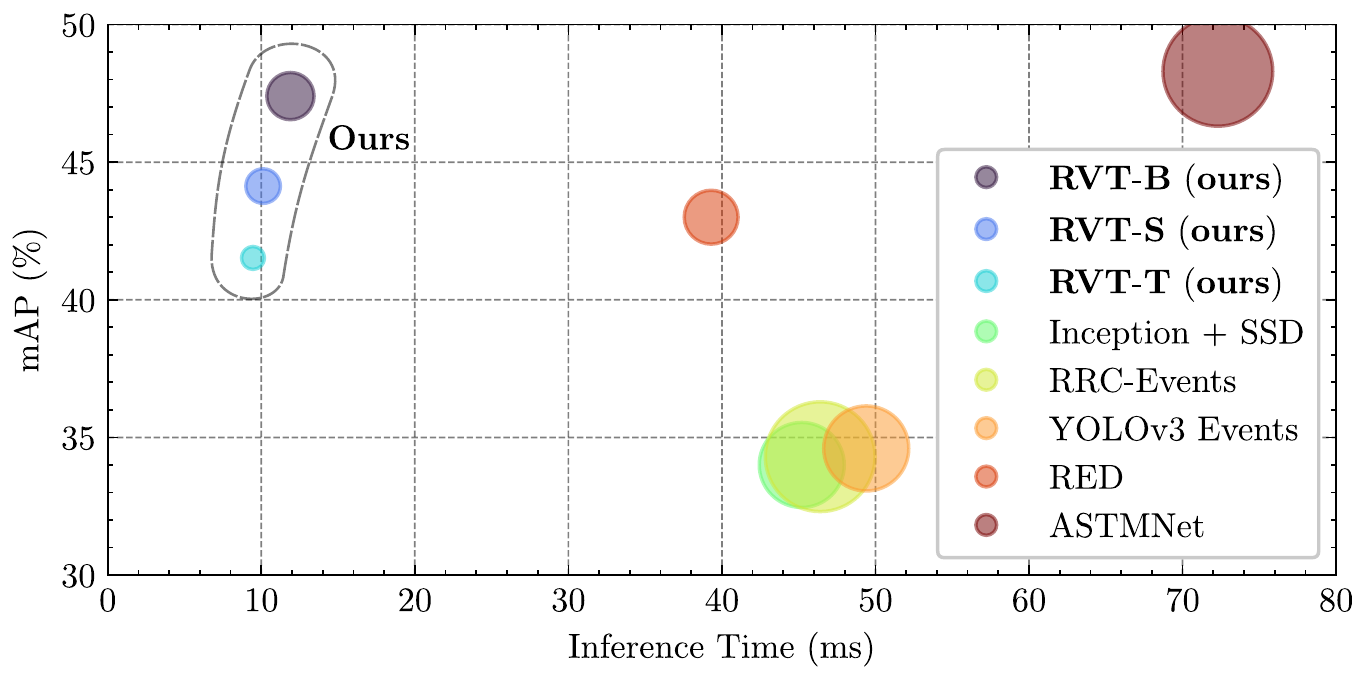}
    \caption{\textbf{Detection performance vs inference time} of our RVT models on the 1 Mpx detection dataset using a T4 GPU. The circle areas are proportional to the model size.}
    \label{fig:first}
\end{figure}

In this work, we aim to utilize these outstanding properties of event cameras for object detection in time-critical scenarios.
Therefore, our objective is to design an approach that reduces the processing latency as much as possible while maintaining high performance.
This is challenging because event cameras asynchronously trigger binary events that are spread of pixel space and time.
Hence, we need to develop detection algorithms that can continuously associate features in the spatio-temporal domain while simultaneously satisfying strict latency requirements.

Recent work has shown that dynamic graph neural networks (GNNs) \cite{Li2021, Schaefer21cvpr} and sparse neural networks \cite{Messikommer2020, Zhang_2022_CVPR, Yao2021TemporalwiseAS, Cordone2022ijcnn} can theoretically achieve low latency inference for event-based object detection.
Yet, to achieve this in practical scenarios they either require specialized hardware or their detection performance needs to be improved.

An alternative thread of research approaches the problem from the view of conventional, dense neural network designs \cite{Iacono2018iros, Chen2018cvprw, Jiang2019icra, Perot2020neurips, Li2022tip}.
These methods show impressive performance on event-based object detection, especially when using temporal recurrence in their architectures \cite{Perot2020neurips, Li2022tip}.
Still, the processing latency of these approaches remains beyond 40 milliseconds such that the low-latency aspect of event cameras cannot be fully leveraged.
This raises the question: How can we achieve both high accuracy and efficiency without requiring specialized hardware?

We notice that common design choices yield a suboptimal trade-off between performance and compute.
For example, prior work uses expensive convolutional LSTM (Conv-LSTM) cells \cite{Shi15nips} extensively in their feature extraction stage \cite{Perot2020neurips, Li2022tip} or relies on heavy backbones such as the VGG architecture \cite{Li2022tip}.
Sparse neural networks instead struggle to model global mixing of features which is crucial to correctly locate and classify large objects in the scene.

To achieve our main objective, we fundamentally revisit the design of vision backbones for event-based object detection.
In particular, we take inspiration from neural network design for conventional frame-based object detection and combine them with ideas that have proven successful in the event-based vision literature.
Our study deliberately focuses on \emph{macro} design of the object detection backbone to identify key components for both high performance and fast inference on GPUs.
The resulting neural network is based on a single block that is repeated four times to form a multi-stage hierarchical backbone that can be used with off-the-shelf detection frameworks.

We identify three key components that enable an excellent trade-off between detection performance and inference time.
First, we find that interleaved local- and global self-attention \cite{Tu2022eccv} is ideally suited to mix both local and global features while offering linear complexity in the input resolution.
Second, this attention mechanism is most effective when preceded by a simple convolution that also downsamples the spatial resolution from the previous stage.
This convolution effectively provides a strong prior about the grid-structure of the pixel array and also acts as a conditional positional embedding for the transformer layers \cite{CPE2021}.
Third, temporal recurrence is paramount to achieve strong detection performance with events.
Differently from prior work, we find that Conv-LSTM cells can be replaced by plain LSTM cells \cite{Hochreiter1997} that operate on each feature separately\footnote{equivalent to $1\times 1$ kernel in a Conv-LSTM cell}.
By doing so, we dramatically reduce the number of parameters and latency but also slightly improve the overall performance.
Our full framework achieves competitive performance and higher efficiency compared to state-of-the-art methods.
Specifically, we reduce parameter count (from 100M to 18.5 M) and inference time (from 72 ms to 12 ms)
up to a factor of 6 compared to prior art \cite{Li2022tip}.
At the same time, we train our networks from scratch, showing that these benefits do not originate from large-scale pretraining.

Our paper can be summarized as follows:
(1) We re-examine predominant design choices in event-based object detection pipelines and reveal a set of key enablers for high performance in event-based object detection.
(2) We propose a simple, composable stage design that unifies the crucial building blocks in a compact way. We build a 4-stage hierarchical backbone that is fast, lightweight and still offers performance comparable to the best reported so far.
(3) We achieve state-of-the-art object detection performance of 47.2\% mAP on the Gen1 detection dataset \cite{gen12020} and a highly competitive 47.4\% mAP on the 1 Mpx detection dataset \cite{Perot2020neurips} while training the proposed architecture from scratch.
In addition, we also provide insights into effective data augmentation techniques that contribute to these results.
\begin{figure*}[ht!]
    \centering
    \includegraphics[width=\linewidth]{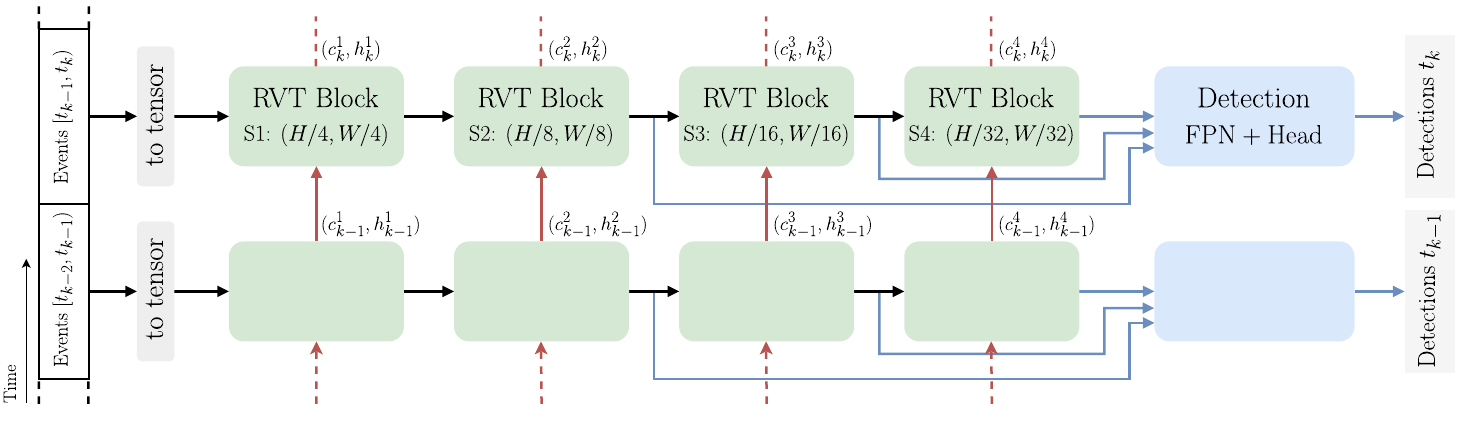}
    \caption{\textbf{Overview} of the unrolled computation graph of our multi-stage recurrent backbone. Events are processed into a tensor representation before they are used as input to the first stage. Each stage also reuses the LSTM states (c: cell, h: hidden) from the previous timestep. Finally, the detection framework interfaces with the backbone from the second stage onwards. Specifically, the hidden states of the LSTMs are used as features for the detection framework.}
    \label{fig:overview}
\end{figure*}
\section{Related Work}\label{sec:relwork}

\subsection{Object Detection for Event Cameras}

Object detection in the event camera literature can be broadly classified into three emerging research directions.

Recent work explores \textbf{graph neural networks} to dynamically construct a spatio-temporal graph \cite{Li2021, Schaefer21cvpr, Mitrokhin_2020_CVPR}.
New nodes and node edges are established by sub-sampling events and finding existing nodes that are close in space-time.
The main challenge is to design the architecture such that information can propagate over vast distances in the space-time volume.
This is relevant, for example, when large objects move slowly with respect to the camera.
Furthermore, aggressive sub-sampling of events can lead to the removal of potentially crucial information, but is often required to maintain low-latency inference.

A second line of work employs \textbf{spiking neural networks} (SNNs) that propagate information sparsely within the network \cite{Zhang_2022_CVPR, Yao2021TemporalwiseAS, Cordone2022ijcnn}.
SNNs are closely related to dense recurrent neural networks (RNNs) in that each spiking neuron has an internal state that is propagated in time.
Differently from RNNs, neurons in SNNs only emit spikes whenever a threshold is reached.
This spike generation mechanism is not differentiable, which leads to substantial difficulties in optimizing these networks \cite{Neftci2019, Shrestha18nips, Lee_2016, Zenke_2018, Lee_2020, Bellec_2020, Taherkhani_2020}.
One workaround is to avoid the aforementioned threshold and instead propagate features throughout the receptive field \cite{Messikommer2020} .
The downside of this mechanism is that the sparse-processing property is lost within deeper layers of the network.
Overall, the design and training of SNNs still requires fundamental investigation before competitive performance can be reached.

A third research direction is concerned with exploring \textbf{dense neural networks} for object detection with event cameras.
The first step is the creation of a dense tensor (event representation) that enables compatibility with dense operations such as convolutions.
Early work directly uses a single event representation generated from a short temporal window of events to infer detections \cite{Iacono2018iros, Chen2018cvprw, Jiang2019icra}.
These approaches discard relevant information from beyond the considered temporal window such that detecting slowly moving objects becomes difficult or impossible.
Followup work addresses this issue by incorporating recurrent neural network layers \cite{Perot2020neurips, Li2022tip} which drastically improved the detection performance.
We follow this line of work but revamp dominant architecture choices to build a canonical framework that is fast, lightweight and highly performant.

\subsection{Vision Transformers for Spatio-Temporal Data}
The success of attention-based models \cite{TF2017neurips} in NLP has inspired the exploration of transformer-based architectures in computer vision \cite{dosovitskiy2021an}.
Attention-based models have recently also been explored in video classification \cite{tong2022videomae, MaskedAutoencodersSpatiotemporal2022, Arnab_2021_ICCV, gberta_2021_ICML} where the models are applied directly to a set of frames.
While these approaches have shown promising results in spatio-temporal modelling, they are optimized for offline processing of stored video data.

In event-based vision, attention-based components have found applications in classification \cite{Sabater_2022_CVPR,Wang2022icip} and image reconstruction \cite{Weng2021}, and monocular depth estimation \cite{recurrent_swin}, but their use in object detection has yet to be investigated.

\begin{figure}
    \centering
    \includegraphics[width=\linewidth]{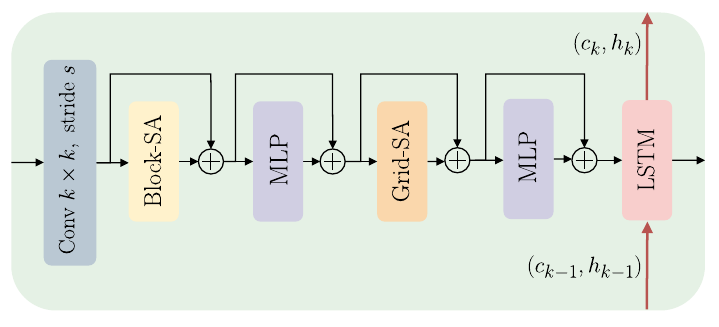}
    \caption{\textbf{RVT block structure}. The input is convolved with kernel size $k\times k$ and stride $s$. Block-SA applies self-attention in local windows while Grid-SA is a global operation using dilated attention. Finally, each block ends with an LSTM that reuses the (cell- and hidden) states from the previous timestep. The LSTM is applied to each feature separately. Normalization and activation layers are omitted for conciseness.}
    \label{fig:block}
\end{figure}

\section{Method}\label{sec:method}
Our object detection approach is designed to process a stream of events sequentially as they arrive.
Incoming events are first processed into tensors that represent events in space and time.
In every timestep, our network takes a new event representation as input as well as the previous states of the recurrent neural network layers.
After each pass through the backbone, the output of the RNNs are used as input to the detection framework.
The following sections elaborate on each one of these steps.
Fig. \ref{fig:overview} shows an overview of the RVT architecture.

\paragraph{Event Processing}\label{sec:ev_repr}
Each pixel of an event camera can independently trigger an event when a significant log brightness change occurs.
An event can be positive or negative depending on the sign of the brightness change.
We characterize an event with polarity $p_k\in\{0,1\}$ as a tuple $e_k=(x_k,y_k,t_k,p_k)$ that occurs at pixel $(x_k, y_k)$ at time $t_k$.
Modern event cameras can produce 10s of millions of events per second which renders event-by-event processing out of reach on conventional processing units.%

In this work we opt for a very simple preprocessing step to enable compatibility with convolutional neural network layers which, as we will show later in Sec. \ref{sec:ablatation}, are an important contributor to the performance of our model.

Our preprocessing step starts with the creation of a 4-dimensional tensor $E$.
The first dimension consists of two components and represents the polarity. The second dimension has $T$ components and is associated with $T$ discretization steps of time. The 3rd and 4th dimension represent height and width of the event camera.
We process set of events $\mathcal{E}$ within a time duration $[t_a,t_b)$ the following way:
\begin{align*}
    E(p, \tau, x, y) &= \sum_{e_k\in\mathcal{E}}\delta(p - p_k)\delta(x-x_k,y-y_k)\delta(\tau-\tau_k),\\
    \tau_k &= \left\lfloor \frac{t_k-t_a}{t_b - t_a}\cdot T\right\rfloor
\end{align*}
In words, we create $T$ 2-channel frames where each pixel contains the number of positive or negative events within one of the $T$ temporal frames.
As a final step, we flatten the polarity and time dimension to retrieve a 3-dimensional tensor with shape $(2T, H, W)$ to directly enable compatibility with 2D convolutions.
We implement the presented algorithm with byte tensors to save memory and bandwidth.
Other, more sophisticated representations are possible \cite{Cannici2019cvprw, Zhu19cvpr, Gehrig19iccv, Baldwin2022, Li2022tip, Tulyakov_2019_ICCV, Barchid2022}, but their thorough evaluation is not our focus.

\subsection{Mixing Spatial and Temporal Features}
The main difficulty of object detection with event cameras is that at any given time, the neural network should be able to efficiently
(1) extract local- and global task-relevant features in pixel space because objects can cover both very small regions or large portions of the field of view;
(2) extract features from very recent events (e.g. moving edges) as well as events from several seconds ago. This is necessary because some objects are moving slowly with respect to the camera such that they generate very few events over time.
These observations motivate us to investigate transformer layers for spatial feature extraction and recurrent neural networks for efficient temporal feature extraction.
Fig. \ref{fig:block} illustrates the components of a single stage.

\paragraph{Spatial Feature Extraction}
The spatial feature extraction stage should incorporate a prior about the fact that pixels are arranged in a 2D grid as early as possible in the computation graph.
We enable this by using a convolution with overlapping kernels on the input features that at the same time spatially downsamples the input or features from the previous stage.
This convolution also endows our model with a conditional positional embedding \cite{CPE2021} such that we do not require absolute \cite{TF2017neurips,dosovitskiy2021an} or relative \cite{liu2021Swin} positional embeddings.
Our ablation study in Sec. \ref{sec:ablatation} shows that overlapping kernels lead to a substantial boost in detection performance.

In a subsequent step, the resulting features are transformed through multi-axis self-attention.
We quickly summarize the steps but refer to Tu et. al \cite{Tu2022eccv} for an elaborate explanation.
Multi-axis attention consists of two stages using self-attention.
The first stage performs local feature interaction while the second stage enables dilated global feature mixing.
More specifically, the features are first grouped locally into non-overlapping windows:
Let $X\in\mathbb{R}^{H\times W\times C}$ be the input feature tensor.
We reshape the tensor to a shape $(\frac{H}{P}\times \frac{W}{P}, P\times P, C)$ where $P\times P$ is the window size in which multi-head self-attention \cite{TF2017neurips} is applied.
This \emph{block attention} (Block-SA in Fig. \ref{fig:block}) is used to model local interactions.
As a next step, we would ideally be able to extract features globally.
One straightforward way to achieve this would be applying self-attention on the whole feature map.
Unfortunately, global self-attention has quadratic complexity in the number of features.
Instead, we use \emph{grid attention} (Grid-SA in Fig. \ref{fig:block}).
Grid attention partitions the feature maps into a grid of shape $(G\times G, \frac{H}{G}\times\frac{W}{G}, C)$ using a $G\times G$ uniform grid.
The resulting windows are of size $\frac{H}{G}\times\frac{W}{G}$.
Self-attention is then applied to these windows which corresponds to global, dilated mixing of features.

We study alternative designs as part of our architecture in the ablation studies in Sec. \ref{sec:ablatation}.

\paragraph{Temporal Feature Extraction}
We opt for temporal feature aggregation with LSTM \cite{Hochreiter1997} cells at the end of the stage.
Differently from prior work \cite{Perot2020neurips, Li2022tip} we find that temporal and spatial feature aggregation can be completely separated.
This means that we use plain LSTM cells such that the states of the LSTMs do not interact with each other.
By avoiding Conv-LSTM units \cite{Shi15nips}, we can drastically reduce the computational complexity and parameter count.
I.e. a Conv-LSTM with kernel size $k\times k$ and stride 1 demands $k^2$ the number of parameters and compute compared to the original LSTM cell.
We examine this aspect in the experimental Sec. \ref{sec:ablatation}.

\paragraph{Model Details}
We apply LayerNorm \cite{Lei16arxiv} before and LayerScale \cite{Touvron_2021_ICCV} after each attention and MLP module, and add a residual connection after each module.
We found that LayerScale enables a wider range of learning rates.

\subsection{Hierarchical Multi-Stage Design}
We compose multiple RVT blocks together to form a multi-stage hierarchical backbone.
The overall architecture is shown in Fig. \ref{fig:overview}.

At first, a local temporal slice of events is processed into a 2D tensor format as formulated in the beginning of this section.
Subsequently, each stage takes the previous features as input and optionally uses the LSTM state from the last timestep to compute features for the next stage.
By saving the LSTM states for the following timestep, each recurrent stage can retain temporal information for the whole feature map.

We follow prior work and use features from the second to the fourth stage for the object detection framework.
To do so, we reshape the hidden states of the LSTMs into 2D feature maps.
\section{Experiments}\label{sec:exp}

\begin{table}[t]
    \centering
    \resizebox{\columnwidth}{!}{%
    \begin{tabular}{@{}lllllll@{}}\toprule
              &        &        &        & \multicolumn{3}{c}{Channels} \\ \cmidrule(l){5-7} 
        Stage & Size   & Kernel & Stride & RVT-T    & RVT-S     & RVT-B        \\ \midrule
        S1    & $1/4$  & 7      & 4      & 32       & 48        & 64           \\
        S2    & $1/8$  & 3      & 2      & 64       & 96        & 128          \\
        S3    & $1/16$ & 3      & 2      & 128      & 192       & 256          \\
        S4    & $1/32$ & 3      & 2      & 256      & 384       & 512          \\ \bottomrule
    \end{tabular}
    }
    \caption{\textbf{RVT parameters and architecture variation}. All model variants share the same parameter set except the number of channels per stage. Each stage initially applies a 2D convolution with kernel and stride as indicated in the table.}
    \label{tab:arch_variants}
\end{table}

We conduct ablations and evaluation our model on the Gen1 \cite{gen12020} and 1 Mpx \cite{Perot2020neurips} event camera datasets.
We train three variants of our model on both datasets: the base model RVT-B and its small and tiny variants RVT-S and RVT-T.
Parameter details for the models are shown in Tab. \ref{tab:arch_variants}.

\subsection{Setup}
\paragraph{Implementation Details}
We initialize all layers randomly except LayerScale which is initialized to 1e-5 for each module.
Our models are trained with mixed precision for 400k iterations with the ADAM optimizer \cite{Kingma15iclr} using a OneCycle learning rate schedule \cite{onecyclelr} with a linear decay from a maximum learning rate.
We use a mixed batching strategy that applies backpropagation through time (BPTT) to half of the samples of the batch and truncated BPTT (TBPTT) to the other half.
More details regarding this batching strategy can be found in the supplementary material.
Our data augmentation includes random horizontal flipping, zooming in and zooming out.
More details on data augmentation are available in Sec. \ref{sec:ablatation} and the supplementary material.
To construct event representations, we consider 50 ms time windows that are discretized into $T=10$ bins.
Finally, we use the YOLOX framework \cite{yolox2021}, which includes the IOU loss, class loss and regression loss.
These losses are averaged both over the batch and sequence length for each optimization step.

To compare against prior work on the Gen1 dataset, we train our models with a batch size of 8, sequence length of 21, and learning rate of 2e-4.
The training takes approximately 2 days on a single A100 GPU.

On the 1 Mpx dataset, we train with a batch size of 24, sequence length of 5, and learning rate of 3.5e-4.
The training takes approximately 3 days on two A100 GPUs.

\paragraph{Datasets}
The Gen1 Automotive Detection dataset \cite{gen12020} consists of 39 hours of event camera recordings at a resolution of $304 \times 240$.
In total, the Gen1 dataset contains 228k car and 28k pedestrian bounding boxes available at 1, 2 or 4 Hz.
We follow the evaluation protocol of prior work \cite{Perot2020neurips, Li2022tip} and remove bounding boxes with a side length of less than 10 pixels and a diagonal of less than 30 pixels.

The 1 MPx dataset \cite{Perot2020neurips} also features driving scenarios but provides recordings at a higher resolution of $720 \times 1280$ over a period of several months at day and night.
It consists of approximately 15 hours of event data labeled at a frequency of 30 or 60 Hz with a total amount of 25 million bounding box labels for three classes (car, pedestrian, and two-wheeler).
We follow the evaluation protocol of prior work \cite{Perot2020neurips, Li2022tip}.
That is, we remove bounding boxes with a side length of less than 20 pixels and a diagonal of less than 60 pixels and halve the input resolution to nHD resolution ($640\times 360$).
We provide qualitative examples of this dataset together with predictions of our base model in Fig. \ref{fig:qual_gen4}.

For both datasets, mean average precision (mAP) is the main metric \cite{Lin14eccv} that we consider.

\subsection{Ablation Studies}\label{sec:ablatation}
This section examines the two main contributors to the final performance of the proposed model.
First, we investigate key components and design choices of the proposed backbone.
Second, we study the influence of different data augmentation techniques that are compatible with our sequential problem setting.

Unless stated otherwise, the ablation studies are performed on the Gen1 validation set using the best performing model after 400k iterations.
To reduce the training time, we use BPTT with a sequence length of 11 instead 21.

\begin{table}[t]
    \centering
    \begin{tabular}{@{}llllll@{}}
        \toprule
                         & \multicolumn{2}{c|}{Gen1}     & \multicolumn{2}{c}{1 Mpx}     &            \\ \cmidrule(lr){2-5}
        Block-type       & mAP           & AP$_{50}$     & mAP           & AP$_{50}$     & Params (M) \\ \midrule
        {\ul multi-axis} & \textbf{47.6} & \textbf{70.1} & \textbf{46.0} & \textbf{72.3} & 18.5       \\
        Swin             & 46.7          & 68.7          & 44.4          & 71.7          & 18.5       \\
        ConvNeXt         & 45.5          & 65.8          & 42.3          & 70.6          & 18.7       \\ \bottomrule
    \end{tabular}
    \caption{\textbf{Spatial Aggregation}. Multi-axis attention leads to the best results on both the Gen1 and 1 Mpx dataset.}
    \label{tab:block_type}
\end{table}

\subsubsection{Model Components}

\paragraph{Spatial Interaction}
In Tab. \ref{tab:block_type}, we study different spatial aggregation techniques.
For a fair comparison, we keep the LSTM and convolutional downsampling layers identical and only exchange the attention and MLP modules.
We compare multi-axis attention with ConvNext blocks \cite{liu2022convnet} and Swin transformer blocks \cite{liu2021Swin}.
ConvNext is a convolutional neural network architecture that has shown competitive performance with transformer-based models on a wide range of tasks, including object detection.
We use the default kernel size of $7 \times 7$ as originally suggested and place three ConvNeXt blocks in each stage to approximately match the number of parameters of the reference model.
Swin, instead, is an attention-based model that applies local self-attention in windows that interact with each other through cyclic shifting.

We find that our Swin variant achieves better performance than the ConvNext variant, however, both are outperformed by multi-axis self-attention \cite{Tu2022eccv} on both the Gen1 and 1 Mpx dataset.
This experiment suggests that global interaction at every stage (multi-axis) is advantageous to purely local interaction (Swin, ConvNext).

\begin{table}[t]
    \centering
    \begin{tabular}{@{}lllll@{}}
        \toprule
        Conv. kernel type      & mAP & AP$_{50}$ & AP$_{75}$ & Params (M) \\ \midrule
        \uline{overlapping}     & \textbf{47.6}    & \textbf{70.1}          & \textbf{52.6}          & 18.5       \\
        non-overlapping & 46.1    & 68.6          & 50.5          & \textbf{17.6}       \\ \bottomrule
    \end{tabular}
    \caption{\textbf{Downsampling Strategy}. The usage of overlapping kernels leads to higher performance at the expense of a slight increase in the number of parameters.}
    \label{tab:patch_type}
\end{table}

\paragraph{Convolutional Downsampling}
The original vision transformer \cite{dosovitskiy2021an} architecture does not perform local feature interaction with convolutional layers.
Some popular hierarchical counterparts also choose to apply downsample features without overlapping kernels \cite{chu2021twins,liu2021Swin}.
In Tab. \ref{tab:patch_type}, we compare overlapping and non-overlapping convolutional kernels in both the input layer (patch embedding) and feature downsampling stage.
While non-overlapping convolutions reduce the number of parameters, they cause a substantial drop in performance.
Consequently, we choose overlapping kernels in all stages of the network.

\begin{table}[t]
    \centering
    \begin{tabular}{@{}lllll@{}}
        \toprule
        LSTM kernel size      & mAP & AP$_{50}$ & AP$_{75}$ & Params (M) \\ \midrule
        \uline{$1\times 1$}           & \textbf{47.6}    & \textbf{70.1}          & \textbf{52.6}          & \textbf{18.5}       \\
        $3\times 3$           & 46.5    & 69.0          & 51.4          & 40.8       \\
        $3\times 3$ depth-sep & 46.3    & 67.2          & 51.2          & 18.6       \\ \bottomrule
    \end{tabular}
    \caption{\textbf{LSTM kernel size}. Conv-LSTM variants do not outperform the feature specific ($1\times 1$) LSTM.}
    \label{tab:lstm_type}
\end{table}

\paragraph{LSTM with Convolutions}
Prior state-of-the-art approaches on object detection with event cameras heavily rely on convolutional LSTM cells \cite{Perot2020neurips,Li2022tip}.
We revisit this design choice and experiment with plain LSTM cells and a depthwise separable Conv-LSTM variant \cite{Pfeuffer2019}.
The depthwise separable Conv-LSTM first applies a depthwise separable convolution on both the input and hidden state before a point-wise ($1\times 1$) convolution is applied.
Our results in Tab. \ref{tab:lstm_type} suggest that plain LSTM cells are sufficient in our model and even outperform both variations.
This is to some degree surprising because both variants are a strict superset of the plain LSTM.
We decide to use a plain LSTM cell based on these observations.

\begin{table}[t]
    \centering
    \begin{tabular}{lllllll}
        \toprule
        S1 & S2    & S3         & S4         & mAP           & AP$_{50}$ & AP$_{75}$ \\ \midrule
           &       &            &            & 32.0          & 54.8      & 31.4      \\
           &       &            & \checkmark & 39.8          & 63.5      & 41.6      \\
           &       & \checkmark & \checkmark & 44.2          & 68.4      & 47.5      \\
                   & \checkmark & \checkmark & \checkmark    & 46.9          & 70.0      & 50.8      \\
        \uline{\checkmark} & \uline{\checkmark} & \uline{\checkmark} & \uline{\checkmark}    & \textbf{47.6} & \textbf{70.1}      & \textbf{52.6}      \\ \bottomrule
    \end{tabular}
    \caption{\textbf{LSTM placement}. LSTM cells contribute to the overall performance even in the early stages.}
    \label{tab:lstm_placement}
\end{table}
\begin{table*}[ht!]
    \centering
    \begin{tabular}{@{}llllllll@{}}
        \toprule
                              &                   &                & \multicolumn{2}{c|}{Gen1}           & \multicolumn{2}{c}{1 Mpx}           &            \\ \cmidrule(lr){4-7}
        Method                & Backbone          & Detection Head & mAP           & Time (ms)           & mAP           & Time (ms)           & Params (M) \\ \midrule
        NVS-S \cite{Li2021iccv}                 & GNN               & YOLOv1 \cite{Redmon2016cvpr}         & 8.6           & -                   & -             & -                   & 0.9        \\
        Asynet \cite{Messikommer2020}                & Sparse CNN        & YOLOv1         & 14.5          & -                   & -             & -                   & 11.4       \\
        AEGNN \cite{Schaefer21cvpr}                & GNN               & YOLOv1         & 16.3          & -                   & -             & -                   & 20.0       \\
        Spiking DenseNet \cite{Cordone2022ijcnn}      & SNN               & SSD \cite{Liu2016eccv}            & 18.9          & -                   & -             & -                   & 8.2        \\
        Inception + SSD \cite{Iacono2018iros}       & CNN               & SSD            & 30.1          & 19.4                & 34.0          & 45.2                & $>60$*      \\
        RRC-Events \cite{Chen2018cvprw}            & CNN               & YOLOv3 \cite{Redmon2018arxiv}         & 30.7          & 21.5                & 34.3          & 46.4                & $> 100$*   \\
        MatrixLSTM \cite{Cannici2019cvprw}           & RNN + CNN               & YOLOv3         & 31.0          & -                   & -             & -                   & 61.5          \\
        YOLOv3 Events \cite{Jiang2019icra}         & CNN               & YOLOv3         & 31.2          & 22.3                & 34.6          & 49.4                & $>60$*      \\
        RED \cite{Perot2020neurips}                   & CNN + RNN         & SSD            & 40.0          & 16.7                & 43.0          & 39.3                & 24.1       \\
        ASTMNet \cite{Li2022tip}              & (T)CNN + RNN      & SSD            & {\ul 46.7}    & 35.6                & \textbf{48.3} & 72.3                & $>100$*    \\
        \textbf{RVT-B (ours)} & Transformer + RNN & YOLOX \cite{yolox2021}          & \textbf{47.2} & \textbf{10.2 (3.7)} & {\ul 47.4}    & \textbf{11.9 (6.1)} & 18.5       \\ \midrule
        \textbf{RVT-S (ours)} & Transformer + RNN & YOLOX          & 46.5          & 9.5 (3.0)           & 44.1             & 10.1 (5.0)           & 9.9        \\
        \textbf{RVT-T (ours)} & Transformer + RNN & YOLOX          & 44.1          & 9.4 (2.3)           & 41.5             & 9.5 (3.5)           & 4.4        \\ \bottomrule
    \end{tabular}
    \caption{\textbf{Comparisons on test sets of Gen1 and 1 Mpx datasets}. Best results in \textbf{bold} and second best {\ul underlined}. Brackets $(\cdot)$ in runtime indicate the inference time with JIT-compiled code using \texttt{torch.compile}. A star $^*$ suggests that this information was not directly available and estimated based on the publications. Runtime is measured in milliseconds for a batch size of 1. We used a T4 GPU for RVT to compare against indicated timings in prior work \cite{Perot2020neurips, Li2022tip} on comparable GPUs (Titan Xp).}%
    \label{tab:sota}
\end{table*}

\paragraph{LSTM Placement}
In this ablation we study the influence of using temporal recurrence only in a subset of stages or not at all.
For all comparisons, we leave the model exactly the same but reset the states of the LSTMs at selected stages in each timestep.
This way, we can simulate the absence of recurrent layers while keeping the number of parameters constant in the comparisons.

The results in Tab. \ref{tab:lstm_placement} suggest that using no recurrence at all leads to a drastic decline of detection performance.
Enabling the LSTMs in each stage, starting from the fourth consistently leads to enhanced performance.
Surprisingly, we find that adding an LSTM to the first stage also leads to improvements, albeit the increase in mAP is not large.
In general, this experiment suggests that the detection framework benefits from features that have been augmented with temporal information.
Based on our observations, we decide to keep the LSTM also in the first stage.

\begin{table}[t]
    \centering
    \begin{tabular}{@{}lll|lll@{}}
        \toprule
        h-flip & zoom-in & zoom-out & mAP & AP$_{50}$ & AP$_{75}$ \\ \midrule
               &         &          & 38.1    & 59.5          & 41.1          \\
        \checkmark      &         &         & 41.6     & 63.5          &  45.5         \\
               & \checkmark       &          & 45.8    & 67.8          & 49.8          \\
               &         & \checkmark        & 44.1    & 65.7          & 48.4          \\
        \uline{\checkmark}      & \uline{\checkmark}       & \uline{\checkmark}        & \textbf{47.6}    & \textbf{70.1}          & \textbf{52.6}          \\ \bottomrule
    \end{tabular}
    \caption{\textbf{Data Augmentation}. Data augmentation consistently improves the results.}
    \label{tab:augm}
\end{table}
\subsubsection{Data Augmentation}

While data augmentation is not directly related to the model itself, it greatly influences the final result as we will illustrate next.
Here, we investigate three data augmentation techniques that are suitable for object detection on spatio-temporal data:
Random (1) horizontal flipping, (2) zoom-in, and (3) zoom-out.

Zoom-in augmentation randomly selects crops that contain at least one full bounding box at the final timestep of the BPTT sequence (i.e. during training).
This crop is then applied to the rest of the sequence before the crops are rescaled to the default resolution.
This procedure ensures that we have at least a single label to compute the loss function while maintaining the same resolution during training.

Zoom-out augmentation resizes the full input to a lower resolution and randomly places the downscaled input in a zero-tensor initialized at the default resolution.
This procedure is then applied in an identical way to the remaining BPTT sequence.

Table \ref{tab:augm} shows that our model is performing poorly if no data augmentation is applied.
Overall, we find that data augmentation is important to combat overfittig not only on the Gen1 sequence but also on the 1 Mpx dataset.
The most effective augmentation is zoom-in, followed by zoom-out and horizontal flipping.
Based on these results, we decide to apply all data augmentation techniques.
We report the specific hyperparameters in the supplementary material.

\begin{figure*}
    \centering
    \includegraphics[width=\linewidth]{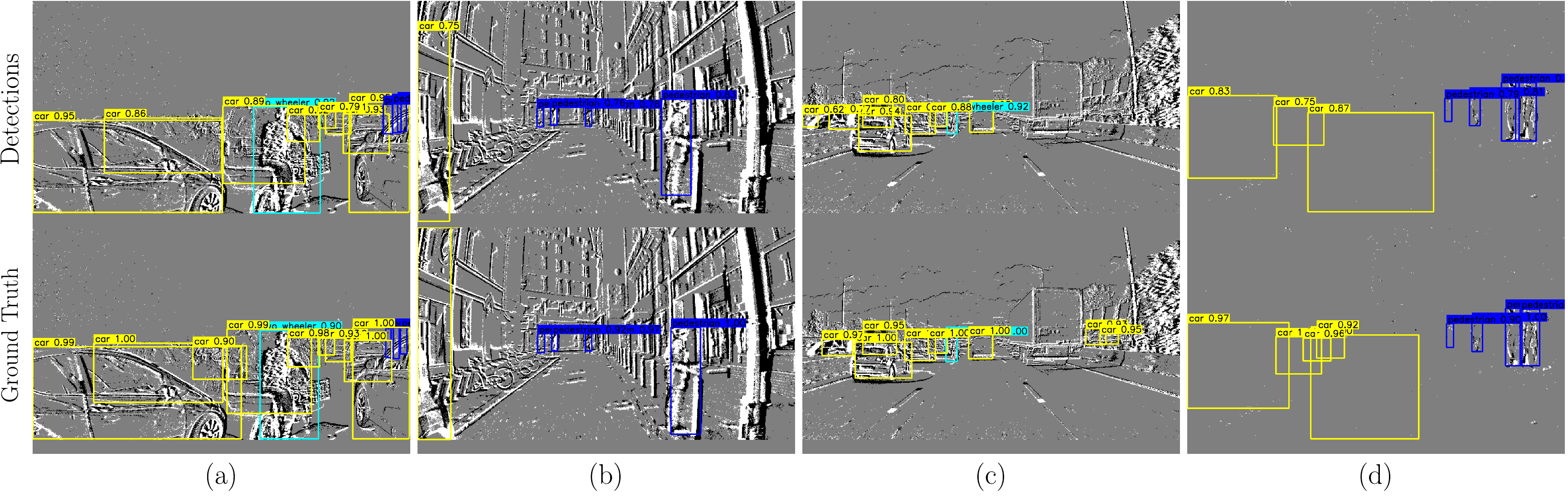}
    \caption{\textbf{Predictions on the 1 Mpx dataset}. All examples are thematically picked to illustrate the behaviour of the model in different scenarios. (d) shows a scenario in which the model can still partially detect objects in absence of events due the temporal memory.}
    \label{fig:qual_gen4}
\vspace*{-10pt}
\end{figure*}

\subsection{Benchmark Comparisons}

In this section, we compare our proposed neural network architecture against prior work on both the Gen1 \cite{gen12020} and 1 Mpx dataset \cite{Perot2020neurips} and summarize the results in Tab. \ref{tab:sota}.
We train three models, a base model (RVT-B) with approximately 18.5 million parameters, a small variant (RVT-S) with 9.9 million parameters, and a tiny model (RVT-T) with 4.4 million parameters by adapting the channel dimensions in each stage.
Their architectural hyperparameters are outlined in Tab. \ref{tab:arch_variants}.
To compare with prior work, we choose the models based on their best performance on the validation set and evaluate them on the test set.

From Tab. \ref{tab:sota} we can draw multiple conclusions.
First, we observe that models using recurrent layers consistently outperform other approaches, both sparse (GNNs, SNNs) and dense feed-forward models without recurrent layers (Inception+SSD, RRC-Events, YOLOv3 Events) by an mAP of more than 10 on both datasets.
One notable exception is MatrixLSTM \cite{Cannici2019cvprw} which applies LSTM cells directly at the input.
In contrast, RED \cite{Perot2020neurips} and ASTMNet \cite{Li2022tip} employ recurrent layers only in deeper layers.

Our base model achieves a new state-of-the-art performance of 47.2 mAP on the Gen1 dataset and 47.4 mAP on the 1 Mpx dataset.
ASTMNet claims comparable results on both datasets albeit at the cost of using a much larger backbone and increased inference time.
The RED model, also reports favorable results, but achieves 7.2 lower mAP on the Gen1 and 4.4 lower mAP on the 1 Mpx dataset compared to our model.
Finally, our tiny model is amongst the smallest in our comparison.
Still, it achieves 4.1 higher mAP on the Gen1 dataset than the RED model while using 5 times fewer parameters.

\paragraph{Inference Time}

We also compute the inference time of our model on a T4 GPU with a batch size of 1.
Unfortunately, both RED and ASTMNet are not open source such that we cannot directly compare inference time on the same GPU model.
Instead, we use the timings provided by the authors that conducted their timing experiments on comparable GPUs (e.g. Titan Xp).
We report the timing results of our models in Tab. \ref{tab:sota} and also visualize them in Fig. \ref{fig:first}.

To compare against prior work we first compute the inference time in PyTorch eager mode.
In eager mode, our base model achieves an inference time of 10.2 milliseconds (ms) on the Gen1 dataset ($304\times 240$ resolution).
This implies a latency reduction of 6 ms compared to RED and over 3 times lower inference time than ASTMNet.
On the the 1 Mpx dataset, at a resolution of $640\times 360$, our base model takes 11.9 ms for a forward pass, which is 3 times faster than RED and over 5 times faster than ASTMNet.

Even on a T4 GPU, most of the inference time is framework overhead.
To overcome this partially, we use the JIT compilation feature \texttt{torch.compile} of PyTorch 2 \cite{pytorch}.
As Tab. \ref{tab:sota} shows, this almost halves the inference time for RVT-B on the 1 Mpx dataset and reduces the inference time by a factor of 2.7 on the Gen1 dataset.
As expected, the small and tiny models benefit even more from JIT compilation.
For example, RVT-T only takes 2.3 ms for a forward pass on Gen1 and 3.5 ms on 1 Mpx.
On a RTX 3090 GPU, RVT-B completes a forward pass in 2.8 ms on the 1 Mpx dataset, which shows the potential for low-latency inference if power consumption is less of a concern.

\section{Discussion and Limitations}\label{sec:limit}
We use a very simple event representation which does not leverage the full potential of event-based data. For example, we only have a weak prior on the order of events because we process the temporal dimension directly with fully connected layers. Recent work has shown substantial gains by introducing temporal convolutions in early layers \cite{Li2022tip}. Efficient low-level processing of event data is still an open research problem that we have not addressed in this work.

Our approach currently only uses event streams to detect objects.
Frames yield complementary information that, when properly incorporated, will yield significantly enhanced detection performance.
For example, in Fig. \ref{fig:qual_gen4} (d) we can see that our model can retain information over some period when no events are available.
Still, the memory of the network will fade and detection performance deteriorates.
High quality frames even at low frame-rate could provide the missing complementary information.
Hence, we believe that a multi-modal extension of our method on a suitable dataset \cite{gehrig2021dsec,Li2019} is a promising next step.
\section{Conclusion}\label{sec:conl}
We introduced a novel backbone architecture for object detection with event cameras.
The architecture consists of a stage design that is repeatedly applied to create a multi-stage hierarchical neural network.
Each stage compactly incorporates convolutional priors, local- and sparse global attention and recurrent feature aggregation.
Our experiments highlight that recurrent vision transformers can be trained from scratch to reach state-of-the-art performance
in object detection with event cameras. 
The resulting canonical stage-design is directly compatible with existing detection frameworks, and paves the way to low-latency object detection with event cameras on conventional hardware.
Nonetheless, we hope that this work also inspires novel designs in future neuromorphic systems.
\section{Acknowledgment}
This work was supported by the National Centre of Competence in Research (NCCR) Robotics (grant agreement No. 51NF40-185543) through the Swiss National Science Foundation (SNSF), and the European Research Council (ERC) under grant agreement No. 864042 (AGILEFLIGHT).

{\small
\bibliographystyle{ieee_fullname}
\bibliography{all}
}

\end{document}


\title{Supplementary Material}

\maketitle

\section{Relationship to Temporal Graph Neural Networks (TGNNs)}
The RVT backbone can be understood as an instance of a discrete-time dynamic graph (DTDG) (section 4.6.1 in \cite{Kazemi_2020}).
A DTDG can be derived from continuous-time dynamic graph (CTDG) by discretizing time.

More specifically, the proposed RVT backbone is an instance of a DTDG with a fixed number of persistent temporal nodes (given a certain input resolution). In our hierarchy, we have four layers of nodes with temporal history (LSTMs in each stage). From here on, there are two different interpretations: (1) The layers up to the LSTMs are a carefully crafted message-passing algorithm interacting directly with the nodes from the previous LSTM nodes closer to the input; or (2) the strided conv layer creates a smaller set of nodes. Block-SA subsequently operates on these nodes by creating a new set of edges (fully graphs connected in local windows) before applying message-passing via self-attention. The same principle applies to Grid-SA which instead creates fully connected graphs in a global dilated grid. The resulting node features are directly used as input to the temporal nodes (LSTM) that also use the states from the previous timestamp to update the node features.

Not every DTDG nor CTDG performs well for event-based vision. In our work, we constrain the design specifically for the problem to get a better trade-off for task performance and inference speed.

\section{Data Pipeline}
\subsection{Mixed Batching Strategy}
Two common ways of training recurrent neural networks is using either backpropagation through time (BPTT) or truncated BPTT (TBPTT).
This section describes a strategy that combines both.

\paragraph{BPTT Dataloading}
For BPTT, the dataloader typically samples short sequences without replacement from the dataset.
As a consequence, the model cannot use an initialized hidden state from earlier parts of the full sequence.
This has the advantage that we can perform more aggressive data augmentation during training.
However, the downside is that the model will fail to generalize to longer sequences.
One possible solution to this problem is training with TBPTT.

\paragraph{TBPTT Dataloading}
For TBPTT, the dataloader again extracts short sequences.
This time, these short sequences are consecutive sequences from a longer sequences that usually cannot be loaded into RAM.
Now we can train the model with initialized, detached hidden states from the previous optimization step.
This allows the model to generalize to longer sequences but also precludes some data augmentation techniques.
For example, zoom-in augmentation on the whole sequence becomes impractical because we could remove too many labels in the process.
Finally, it can also lead to training instabilities and overfitting because the model is optimized on samples from the same sequences for many training steps.

\begin{table}[t]
    \centering
    \begin{tabular}{@{}llll@{}}
        \toprule
        Dataloading Strategy      & mAP & AP$_{50}$ & AP$_{75}$ \\ \midrule
        BPTT            & 38.8    & 66.0          & 39.5 \\
        TBPTT           & 44.1    & 72.0          & 46.1          \\
        Mixed           & \textbf{46.0}    & \textbf{72.3}          & \textbf{49.4}          \\ \bottomrule
    \end{tabular}
    \caption{\textbf{Dataloading Strategy}. Combining BPTT and TBPTT (Mixed) yields the best results on the 1 Mpx validation set.}
    \label{tab:ablation_data}
\end{table}

\paragraph{Mixing BPTT and TBPTT}
What we found to work best is to create two dataloaders:
First, a dataloader that randomly samples short sequences from the dataset.
This is useful for training with BPTT and improves the diversity of samples in the batch.
This dataloader also applies the full set of data augmentations.
Second, a dataloader that collates data from iterating through whole sequences.
This dataloader is used for training with TBPTT and improves the capability of the model to generalize to sequences longer than the training sequence length.
In this case, we do not apply zoom-in augmentation.

Table \ref{tab:ablation_data} compares the performance of the RVT-B model on the validation set of the 1 Mpx dataset using the different dataloading strategies.
The performance of BPTT can be improved by increasing the sequence length at the cost of increased memory consumption and training time.
Instead, the mixed dataloading strategy allows training with a short sequence length (here 5) while enjoying stable training with improved detection performance.

In practice, on each GPU, half of the batch is created with the BPTT dataloader and the other half with the TBPTT dataloader.
Both batches are separately loaded onto the GPU before being collated and fed to the model.
The model then dynamically resets the hidden states based on the information whether each sample in the batch originates from the BPTT or TBPTT dataloader.

\subsection{Data Augmentation Details}\label{sec:data_augm}
We apply three data augmentation techniques to train our models from scratch.
Table \ref{tab:augm} summarizes the probability of each augmentation being used on an individual sample.

For each sample, we may apply horizontal flipping and also apply a zoom augmentation subsequently.
For the zoom augmentation we draw from a Bernoulli distribution to indicate whether we apply the augmentation at all.
If zoom augmentation shall be applied, we randomly choose between zoom-in or zoom-out augmentation based on the respective probability.
For the zoom augmentations, there is also the parameter that defines the \emph{magnitude} with which the augmentation is applied.
A magnitude of 1 means that no zoom is applied while a magnitude greater than 1 indicates how strongly zoom-in or zoom-out is applied with respect to the original resolution.
The magnitude that we finally apply is drawn from a continuous uniform distribution with bounds \emph{min} and \emph{max}.
\begin{table}[t]
    \centering
    \begin{tabular}{@{}llll@{}}
        \toprule
                        &              & \multicolumn{2}{c}{Magnitude} \\ \cmidrule(l){3-4} 
        Augmentation    & Probability  & min           & max           \\ \midrule
        horizontal flip & 0.5          &               &               \\
        apply zoom      & 0.8          &               &               \\
        zoom-in         & 0.8          & 1             & 1.5           \\
        zoom-out        & 1-P(zoom-in) & 1             & 1.2           \\ \bottomrule
    \end{tabular}
    \caption{\textbf{Data Augmentation Parameters}. The probability defines the Bernoulli distribution from which we draw the decision whether to apply this augmentation on a given sample.}
    \label{tab:augm}
\end{table}
\section{Additional Model Details}
The attention window size for all stages is set to $8\times 10$ for the Gen1 dataset and $6\times 10$ for the 1 Mpx dataset. Consequently, the fourth stage employs global self-attention on the Gen1 dataset, whereas it utilizes four windows at the same stage for the 1 Mpx dataset.

\section{Additional Experiments}\label{sec:add_exp}
This section provides two additional experiments that did not fit into the main paper.
First, we briefly discuss an ablation on the possibility of using residual LSTM layers.
Second, we follow with a qualitative study of cross-dataset generalization using our model trained on the 1 Mpx dataset.

\subsection{Residual LSTM Ablation}\label{sec:add_exp_lstm}
Our model employs LSTM cells \cite{Hochreiter1997} without skip/residual connections in the model.
We also experimented with adding skip connections to the LSTM cells on the Gen1 \cite{gen12020} dataset.
Table \ref{tab:lstm_residual} shows that adding a residual connection to the LSTM cells leads to worse results.
We hypothesize that this residual connection hampers the LSTM's ability to control the mixture of incoming (current timestep) and retained temporal features (previous timesteps).
For example, it would be difficult for the residual-LSTM combination to ignore the incoming feature because the output of the LSTM is simply added to this feature.
Without the residual connection, the LSTM could set the input gate to 0 to ignore the input.

\begin{table}[t]
    \centering
    \begin{tabular}{@{}llll@{}}
        \toprule
        LSTM residual & mAP & AP$_{50}$ & AP$_{75}$ \\ \midrule
        \underline{without residual}     & \textbf{47.6}    & \textbf{70.1}          & \textbf{52.6} \\
        with residual & 46.0    & 69.8          & 50.3          \\ \bottomrule
    \end{tabular}
    \caption{\textbf{LSTM with and without residual connection}. Using a skip connection over the LSTM cells leads to worse results on the Gen1 validation set.}
    \label{tab:lstm_residual}
\end{table}
\subsection{Cross-Dataset Generalization: From 1 Mpx to DSEC}\label{sec:add_exp_dsec}
DSEC is a dataset that features event cameras and global shutter cameras close to each other.
Unlike the 1 Mpx dataset \cite{Perot2020neurips} which was recorded mostly urban scenarios in Paris, DSEC \cite{gehrig2021dsec} provides recordings from urban and rural regions in Switzerland.
Furthermore, the event camera used in the DSEC dataset is a Gen 3 prophesee event camera instead of a Gen 4 camera as in the 1 Mpx dataset.
In this section, we qualitatively show that our model can be successfully deployed in a different environment and using different event cameras.

We deploy RVT-B, trained on the 1 Mpx dataset , on several sequences of the DSEC dataset to qualitatively assess the cross-dataset generalization.
While DSEC does not yet provide object detection labels, we can visually assess the quality of the detections by using the provided calibration files to map the frames of the global shutter camera to the event camera view.

\begin{figure*}[ht]
    \includegraphics[width=0.33\linewidth]{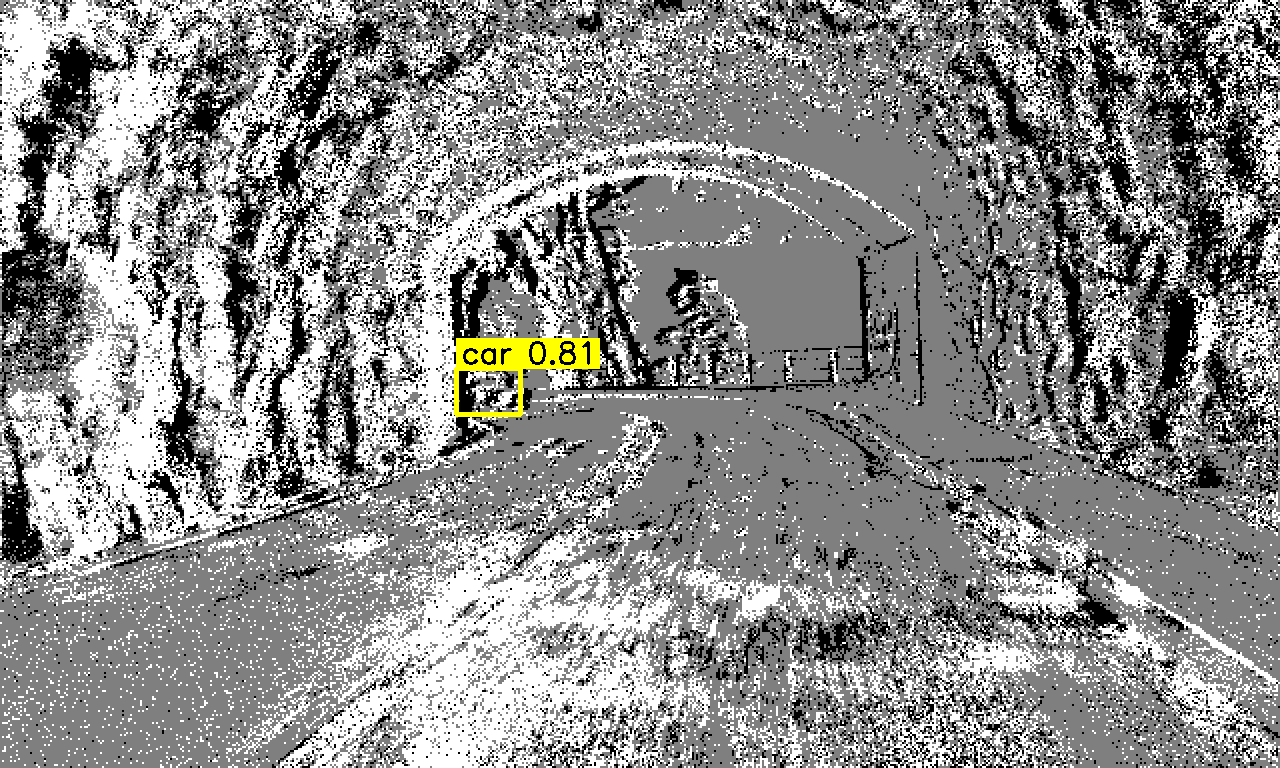}
    \includegraphics[width=0.33\linewidth]{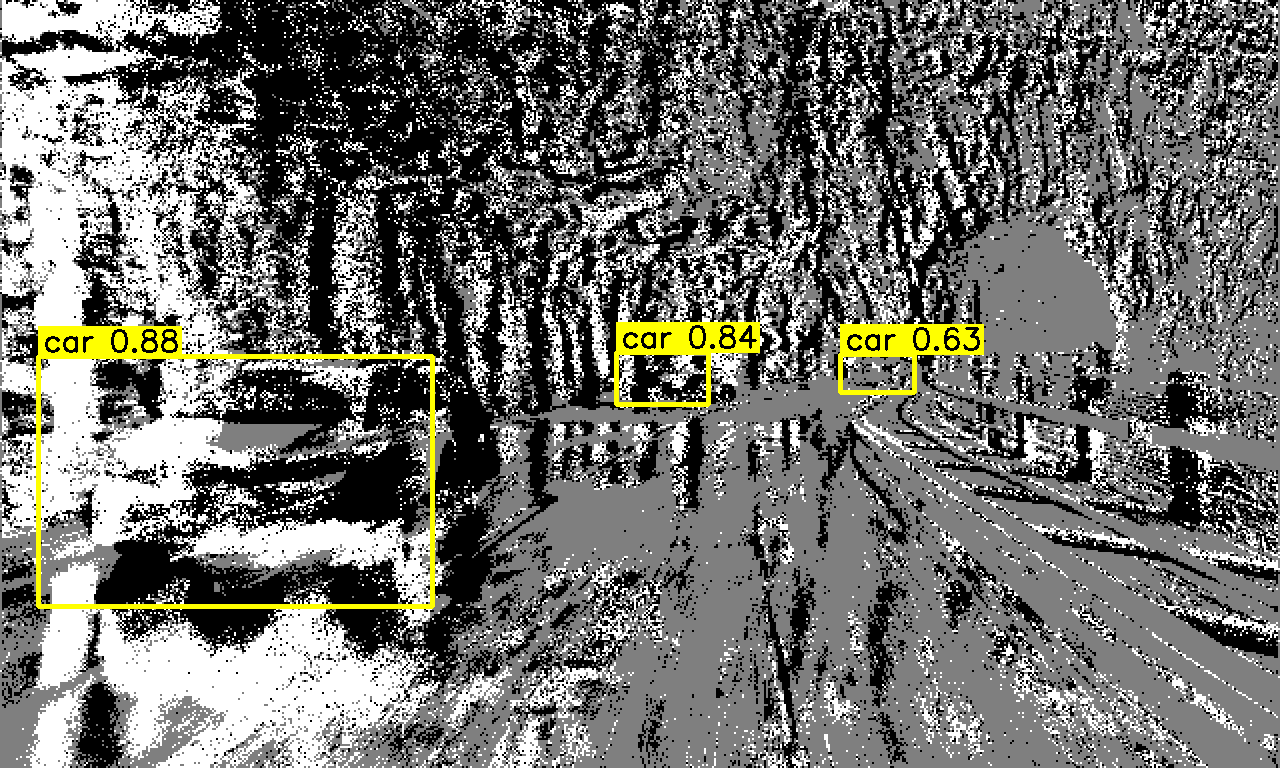}
    \includegraphics[width=0.33\linewidth]{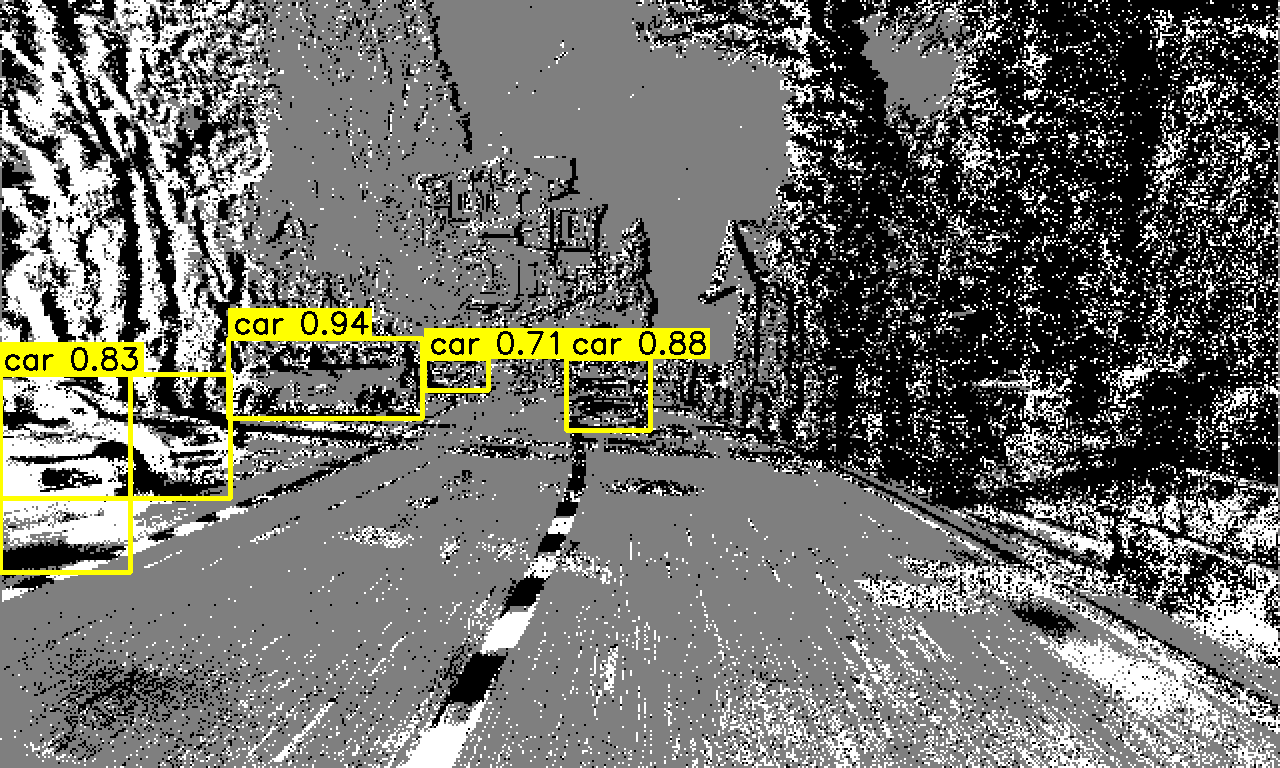}\\
    \includegraphics[width=0.33\linewidth]{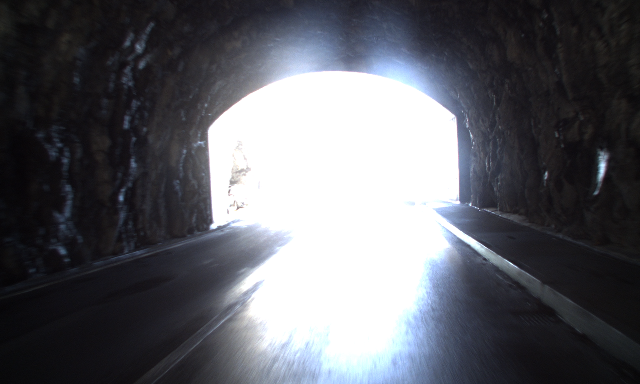}
    \includegraphics[width=0.33\linewidth]{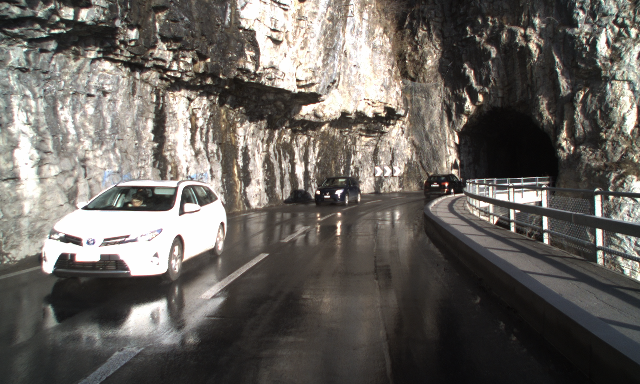}
    \includegraphics[width=0.33\linewidth]{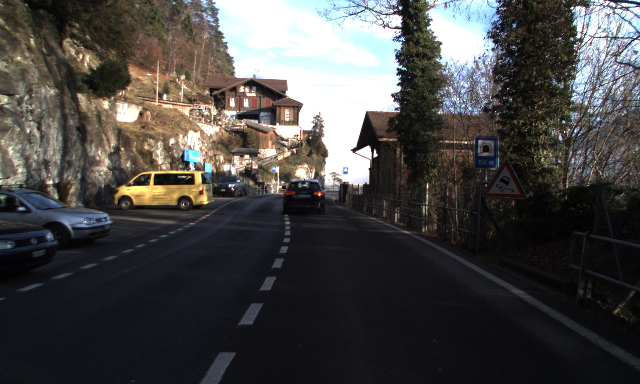}\\
    \includegraphics[width=0.33\linewidth]{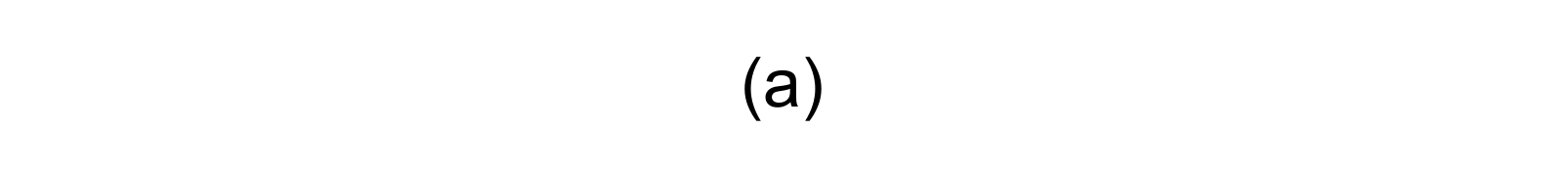}
    \includegraphics[width=0.33\linewidth]{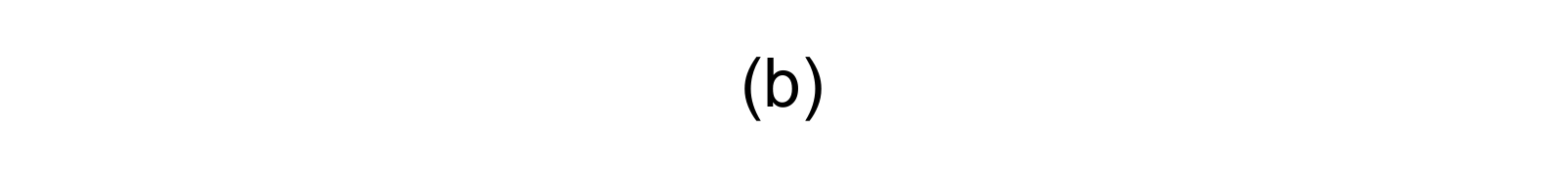}
    \includegraphics[width=0.33\linewidth]{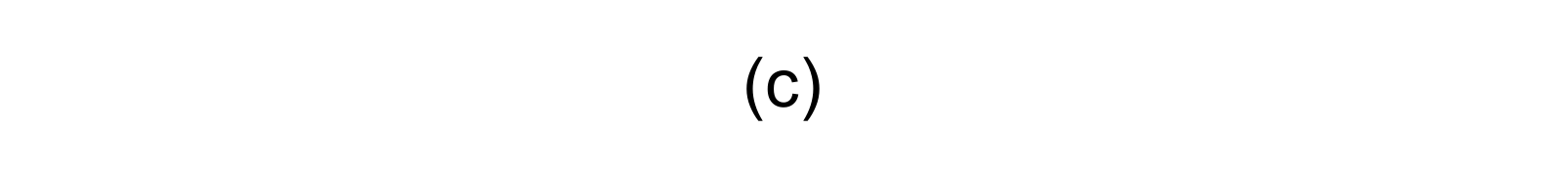}
    \caption{Prediction examples on the DSEC dataset featuring a \textbf{mountainous environment}. Frames are shown only for visualization purposes and are not used by the model. Column (a) shows a typical high-dynamic range (HDR) scenario where the vehicle is exiting a tunnel with a car approaching from outside the tunnel. The HDR capabilities of the event cameras enables our model to detect the approaching car. Column (b) shows a scenario with a wet road and challenging reflections.}
    \label{fig:dsec_interlaken}
\end{figure*}

\begin{figure*}[ht]
    \includegraphics[width=0.33\linewidth]{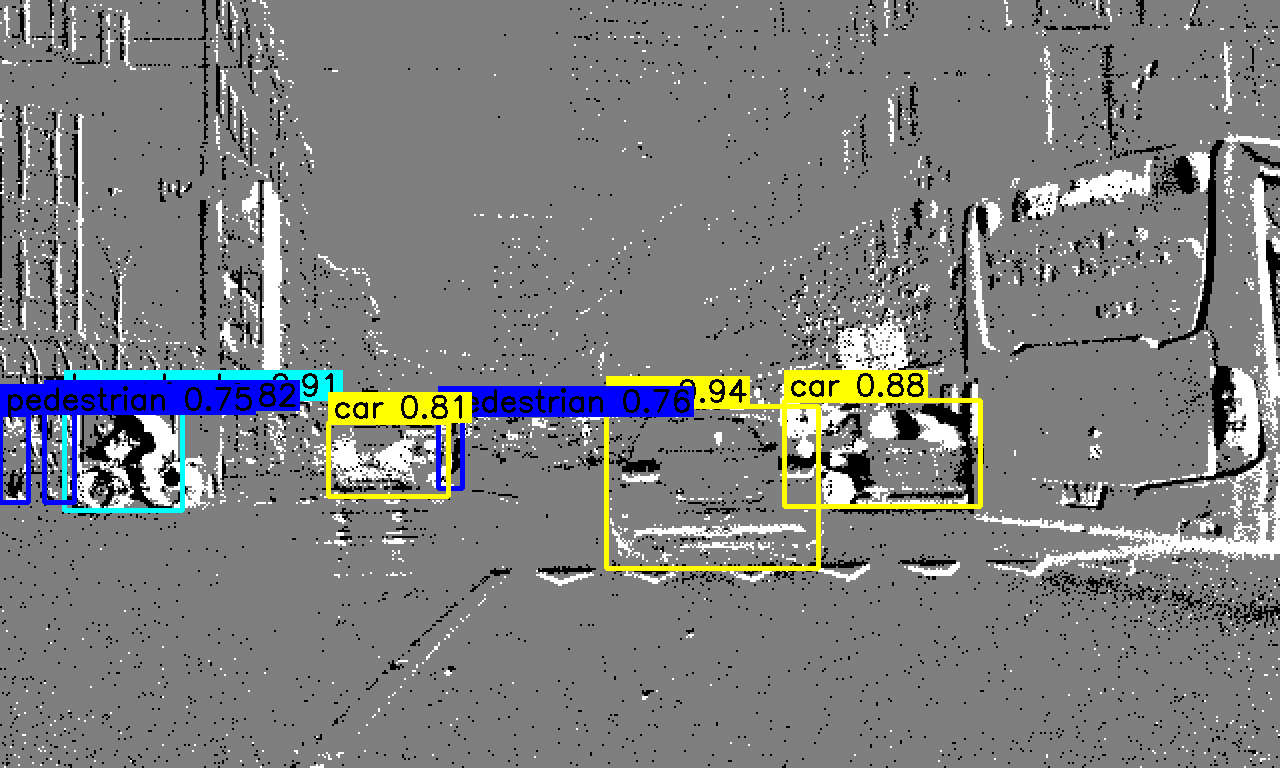}
    \includegraphics[width=0.33\linewidth]{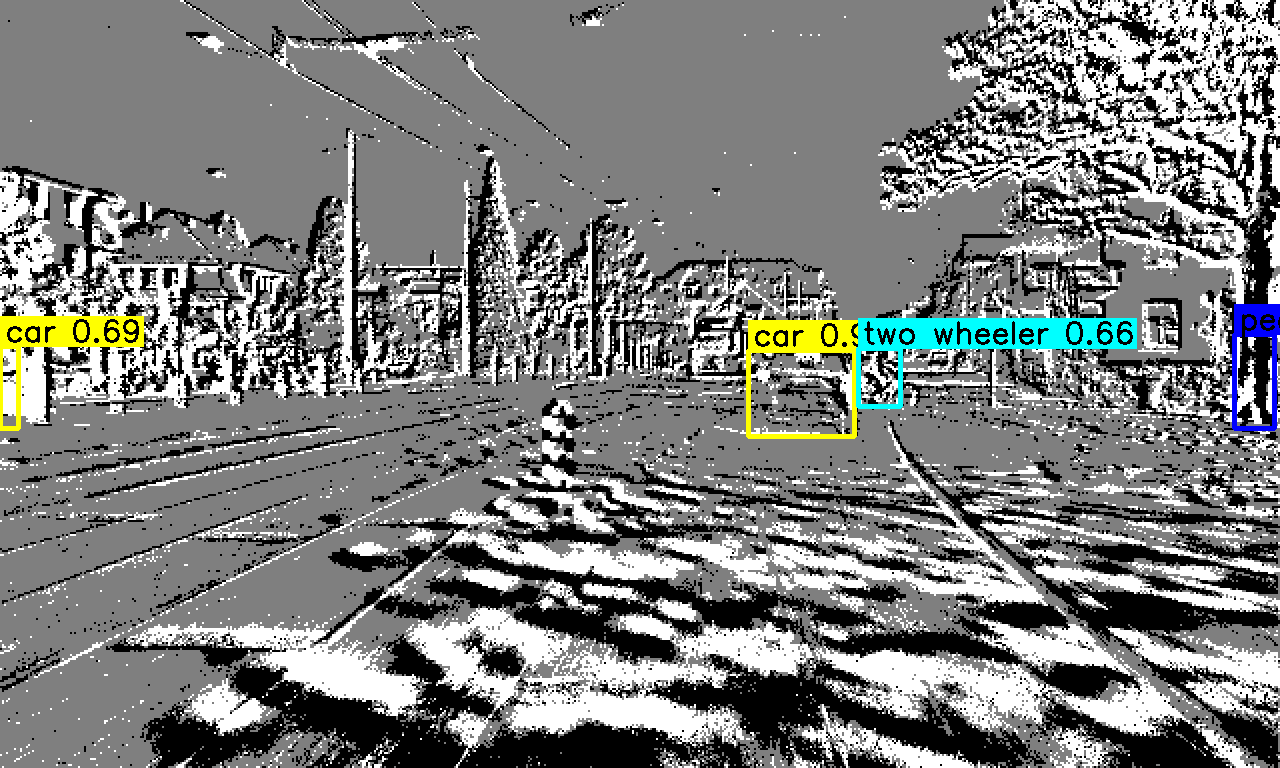}
    \includegraphics[width=0.33\linewidth]{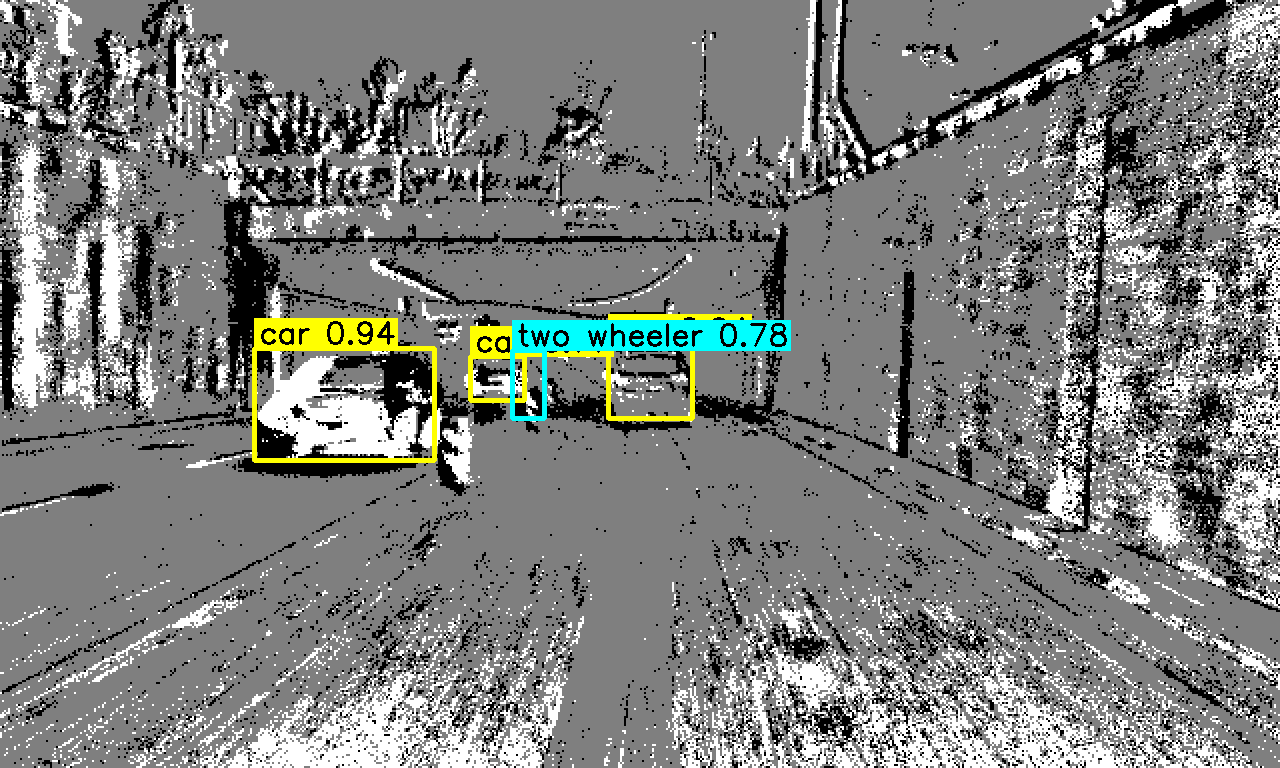}\\
    \includegraphics[width=0.33\linewidth]{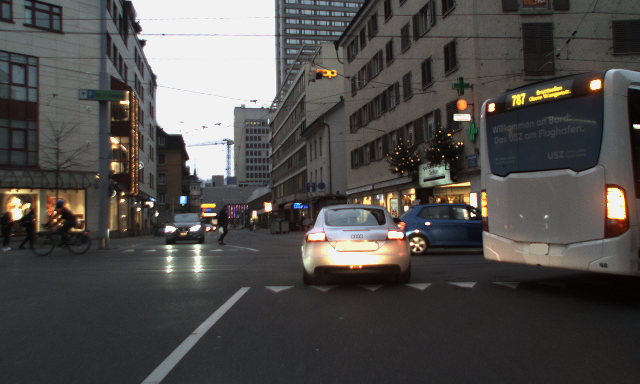}
    \includegraphics[width=0.33\linewidth]{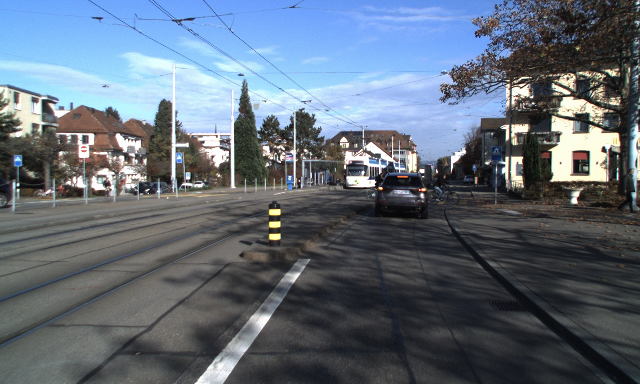}
    \includegraphics[width=0.33\linewidth]{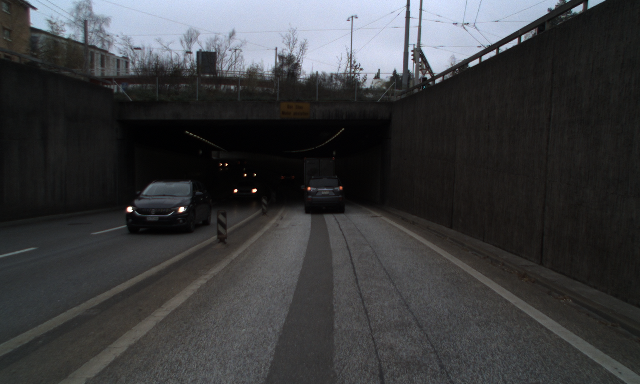}\\
    \includegraphics[width=0.33\linewidth]{imgs/supp/a_subcap_hack.pdf}
    \includegraphics[width=0.33\linewidth]{imgs/supp/b_subcap_hack.pdf}
    \includegraphics[width=0.33\linewidth]{imgs/supp/c_subcap_hack.pdf}
    \caption{Prediction examples on the DSEC dataset featuring a \textbf{(sub-)urban environments}. Frames are shown only for visualization purposes and are not used by the model. Column (a) illustrates a typical urban situation where pedestrians, two-wheelers, cars and other road users occupy the street simultaneously. In column (c), we show a failure case of our model where a street pillar is erroneously detected and classified as a two-wheeler.}
    \label{fig:dsec_zurich}
\end{figure*}

Figure \ref{fig:dsec_interlaken} and \ref{fig:dsec_zurich} show predictions of our model together with the images closest in time to the detections.
Figure \ref{fig:dsec_interlaken} shows our model can successfully detect cars in mountainous environments.
In particular, Figure \ref{fig:dsec_interlaken} (a) shows an HDR scene where the global shutter frame is overexposed such that the approaching car is barely visible.
Due to the high dynamic range of the event camera, our model can detect the approaching car without any difficulty.
Figure \ref{fig:dsec_zurich} features more urban environments where our model also manages to detect objects correctly.

\paragraph{Discussion of Failure Cases}
By and large, our model can successfully detect objects on DSEC even though it has only been trained on the 1 Mpx dataset.
Still, we found failure cases that might stem from distribution shift between the datasets.
For example, Figure \ref{fig:dsec_zurich} (c) shows the erroneous detection of a two-wheeler that instead is a pillar on the street.

Overall, the model is good at detecting cars but is less confident and accurate at detecting two-wheelers and pedestrians.
This effect likely stems from the fact that the 1 Mpx dataset has almost twice as many car labels as pedestrian and two-wheeler labels combined. 
\section{Dataset Licenses}
\paragraph{Gen1 \cite{gen12020}}
``Prophesee Gen1 Automotive Detection Dataset License Terms and Conditions'': \url{https://www.prophesee.ai/2020/01/24/prophesee-gen1-automotive-detection-dataset/}
\paragraph{1 Mpx \cite{Perot2020neurips}}
``Prophesee 1MegaPixel Automotive Detection Dataset License Terms and Conditions'': \url{https://www.prophesee.ai/2020/11/24/automotive-megapixel-event-based-dataset/}
\paragraph{DSEC \cite{gehrig2021dsec}}
``Creative Commons Attribution-ShareAlike 4.0 International public license (CC BY-SA 4.0)'': \url{https://dsec.ifi.uzh.ch/}
\balance

{\small
\bibliographystyle{ieee_fullname}
\bibliography{all}
}